\ifdefined\XeTeXversion\else\ifdefined\directlua\else\pdfoutput=1\fi\fi 
\documentclass[11pt]{article}
\usepackage[nologo]{thefinai}
\usepackage{needspace}
\usepackage{placeins}
\title{Human-Oriented World Model} 
\date{\today}  
\author{Saihui Hou\textsuperscript{1}, Zengyun Mou\textsuperscript{1},
  Chunshui Cao\textsuperscript{2}, Yongzhen Huang\textsuperscript{1,2}} 
\affiliation[1]{Beijing Normal University}
\affiliation[2]{WATRIX AI}

\begin{document}

\begin{abstract}
Supporting a person over repeated interactions requires interpreting their current state in context and anticipating its subsequent evolution.
We formulate the \emph{Human-Oriented World Model} (HOWM) as a conceptual framework centered on task-relevant human states and their evolution.
HOWM connects current observations and individual history with environmental and social context to organize continuing state estimation, personalized dynamics prediction, and action support.
Dynamic expectations and new observations jointly inform the current estimate, which in turn anchors personalized forecasts based on shared knowledge and individual experience.
An external policy relates forecasts under alternative conditions to the person's goals, preferences, and constraints when considering support.
Actual responses and outcomes provide evidence for revising the state estimate and personal dynamic patterns, thereby informing subsequent estimation, prediction, and support.
We illustrate the framework through hypothetical training and learning scenarios.
Our research agenda asks whether accumulated interaction experience improves subsequent estimation and prediction, and whether using the model helps the person pursue their goals over time.
\end{abstract}

\maketitle

{\small\sffamily\bfseries\color{finaiblue}
\noindent\emph{Note.
This report presents a conceptual framework and research agenda
for human-oriented world modeling.
Feedback and discussion from the community are welcome.}
\par}
\medskip

\section{Introduction}
\label{sec:introduction}

Understanding a person's evolving state requires relating behavior to individual experience and context.
Consider an assistant following a participant across badminton practice sessions.
An unstable movement may be typical for that participant, reflect a more demanding exercise, or signal a change in their current condition.
Earlier sessions help interpret the current movement and suggest how this participant tends to adjust as an exercise continues.
Supporting the participant's goals also requires anticipating how their condition and performance may develop under different practice arrangements.

World models and human state modeling provide foundations for connecting state representations to predictions of change and interaction~\citep{ha2018worldmodels,fei2026mwm,ding2026combodied}.
Longitudinal memory makes individual experience available across encounters~\citep{zhong2024memorybank,yang2025egolife}.
Across repeated encounters, observations are often incomplete, people differ in their typical behavior and patterns of change, and task or social conditions can alter what an observation means.
Support can itself change both the person's subsequent state and the experience available for later learning.
These dependencies motivate a continuing account that connects each encounter to previous expectations and uses new evidence to revise both the current estimate and the basis for future prediction.

We formulate the \emph{Human-Oriented World Model} (HOWM) as a conceptual framework centered on task-relevant human states and their evolution.
HOWM connects current observations and individual history with environmental and social context to organize continuing state estimation, personalized dynamics prediction, and action support, while actual interaction provides evidence for updating its understanding of the person.
The task determines which aspects of the person and which changes are modeled.
The person's goals, preferences, and constraints guide how proposed support is evaluated.

Estimation and prediction support one another in this process.
An earlier forecast supplies an expectation about the present; new observations and context revise that expectation, and the resulting estimate anchors subsequent forecasts.
An external policy relates these forecasts to the person's goals when proposing support, including proactive offers within an agreed scope of assistance.
Actual responses and outcomes then provide evidence for revising the current estimate and learning personal patterns of change.
For example, repeated practice trajectories may show that a participant adjusts to a change in cue pace sooner than anticipated under comparable conditions.
Revising that pattern can change the next state estimate, forecasts of subsequent performance, and the support considered.

Three capability questions guide the report:
\begin{itemize}
    \item How can current observations, individual history, and context support continuing state estimation?
    \item How can shared knowledge and individual experience support predictions of a person's subsequent development under specified conditions?
    \item How can these forecasts inform support that reflects the person's goals, preferences, and choices?
\end{itemize}

Learning from actual interaction runs across all three: its value depends on whether accumulated experience improves subsequent estimation and prediction, and whether using that understanding helps the person pursue their goals.

Our contributions are twofold:
\begin{itemize}
    \item We organize HOWM's modeling objects and information roles, making explicit how updates to personal dynamics inform subsequent state priors and forecasts.
    \item We specify comparison designs and longitudinal validation pathways that distinguish the contributions of personal baselines, temporal patterns, and continued adaptation, with separate assessment of assistance outcomes in relation to the person's goals.
\end{itemize}


\section{Related Work}
\label{sec:related_work}

\subsection{World Models and Human State Modeling}
\label{sec:related_state}

World models provide a basis for connecting representations of the present, predictions of change, and the use of those predictions in action.
World Models and Dreamer illustrate learning policies through trajectories generated by learned environment models~\citep{ha2018worldmodels,hafner2025dreamer}.
For human modeling, state selection also concerns aspects of the person that help explain and predict behavior.
Machine Theory of Mind uses behavioral observations to infer agent characteristics and mental states and to predict behavior, with experiments on artificial agents in simple gridworld environments~\citep{rabinowitz2018tom}.
Mental World Modeling (MWM) represents a joint physical and mental state, models the target's perspective, and simulates coupled state changes under candidate actions~\citep{fei2026mwm}.

Human perception supplies complementary evidence about the person being modeled.
Identity cues, including gait recognition~\citep{fan2025opengait}, can help associate observations across encounters when identification is reliable; registration can also establish this link.
Pose estimation~\citep{cao2017pose} and annotations of actions localized to people, such as AVA~\citep{gu2018ava}, describe body configuration and activity.
Using such evidence to infer a person's state requires relating behavior to task demands, environmental conditions, and interactions with other people.

Continuing state estimation connects these observations to an account that can be maintained as the person changes.
Bayesian filtering combines a prediction from a transition model with new observational evidence to revise the state estimate~\citep{sarkka2023filtering}.
HOWM builds on this relationship: dynamic expectations inform current state estimates, and revised estimates anchor subsequent personalized forecasts.
For estimates maintained across encounters, the expectations and evidence also depend on individual experience and the conditions under which it was obtained.

\subsection{Longitudinal Memory and Personalized Dynamics}
\label{sec:related_memory_dynamics}

Longitudinal memory provides ways to retain, organize, and retrieve experience for later use.
Generative Agents uses retrieved experiences and higher-level reflections to support simulated behavior~\citep{park2023generativeagents}, while MemoryBank combines retrieval with continuing memory updates and selective forgetting in sustained dialogue~\citep{zhong2024memorybank}.
In long-duration visual observation, EgoLife's EgoButler system combines identity-aware video interpretation through EgoGPT with hierarchical retrieval through EgoRAG to answer questions about recorded daily activities~\citep{yang2025egolife}.
LongMemEval examines memory capabilities in conversational assistants, including temporal reasoning and knowledge updates, and analyzes indexing, retrieval, and reading as distinct stages~\citep{wu2025longmemeval}.
For human state modeling, the relevance of a retrieved episode depends on the starting condition, task, and interaction associated with it.

Personalized dynamics uses such histories to relate shared regularities to a person's typical level and patterns of change.
Random-effects models for longitudinal data provide a statistical basis for modeling repeated observations with parameters that vary across individuals within a shared population model~\citep{laird1982random}.
The distinction between stable differences across people and within-person change also matters for interpreting temporal relationships, as illustrated by \citet{hamaker2015critique} in cross-lagged panel models.
As a domain-specific example of prediction from individual histories, Deep Knowledge Tracing uses a student's interaction sequence to predict subsequent performance~\citep{piech2015dkt}.
These foundations make both a person's typical level and the course of change relevant to personalization.
Applying learned patterns over repeated encounters further raises the question of when new experience warrants updating a pattern or reconsidering where it applies, with task and interaction conditions needed to interpret the observed trajectories.
Memory maintenance, state estimation, and learning personal dynamics have related roles: they determine what evidence is retained, what is believed about the present, and how changes are anticipated.

\subsection{Prediction and Learning in Human-Agent Interaction}
\label{sec:related_interaction}

Action-conditioned prediction describes a person's possible subsequent behavior and states under specified action conditions.
Human--robot interaction provides a concrete example: \citet{sadigh2016planning} model how an autonomous car's actions affect a human driver's responses and incorporate those responses into the dynamics used for planning.
For sustained personal support, the Personal World Model (PWM) within ComBodied Agents is formulated as an individual predictor that uses event history and context to estimate futures under candidate user decisions, agent support, and environmental changes~\citep{ding2026combodied}.
It predicts personal states, events, and outcomes, while an external policy selects support.
HOWM adopts this division: the model supplies conditional futures, and an external policy evaluates support against the person's goals, preferences, and contextual constraints.

Interaction supplies evidence for learning as well as outcomes relevant to the person.
The broader ComBodied Agents framework connects perception, longitudinal memory, personal prediction, and intervention through repeated feedback and distinguishes model quality from human outcomes~\citep{ding2026combodied}.
Building on these foundations, HOWM examines how updates to personal dynamics inform subsequent state priors and forecasts.
Its research agenda develops comparisons that distinguish the contributions of personal baselines, temporal patterns, and continued adaptation.
Assessing assistance additionally follows the person's choices and outcomes relative to their goals.
Micro-randomized trials in mobile health provide a basis for evaluating support, using repeated randomization to estimate proximal effects of intervention components and how those effects vary with time and context~\citep{klasnja2015microrandomized}.
These perspectives motivate longitudinal assessment of model use together with separate evaluation of the effects of support.

\section{HOWM: Conceptual Framework and Modeling Mechanisms}
\label{sec:framework}

\subsection{Working Definition, Task Scope, and Overall Organization}
\label{sec:framework_scope}

We define a \emph{Human-Oriented World Model} (HOWM) as a conceptual framework centered on task-relevant human states and their evolution.
It connects current observations and individual history with environmental and social context to support continuing state estimation, personalized dynamics prediction, and action support, while actual interaction supplies evidence for updating its understanding of the person.

The modeled states are selected by asking which aspects of the person help explain behavior, anticipate relevant developments, and consider support in relation to the person's goals.
Depending on the task, these may concern movement, behavior, cognition, emotion, or goals themselves.
In a training setting, movement stability and reports of effort can supply evidence about current fatigue or readiness; the model then considers how that condition may develop during another exercise or a rest period.
Those possible changes inform choices about the next activity and its demands.
The task scope therefore connects observations to state variables, changes of interest, and support needs, including the environmental conditions and social interactions that shape them.

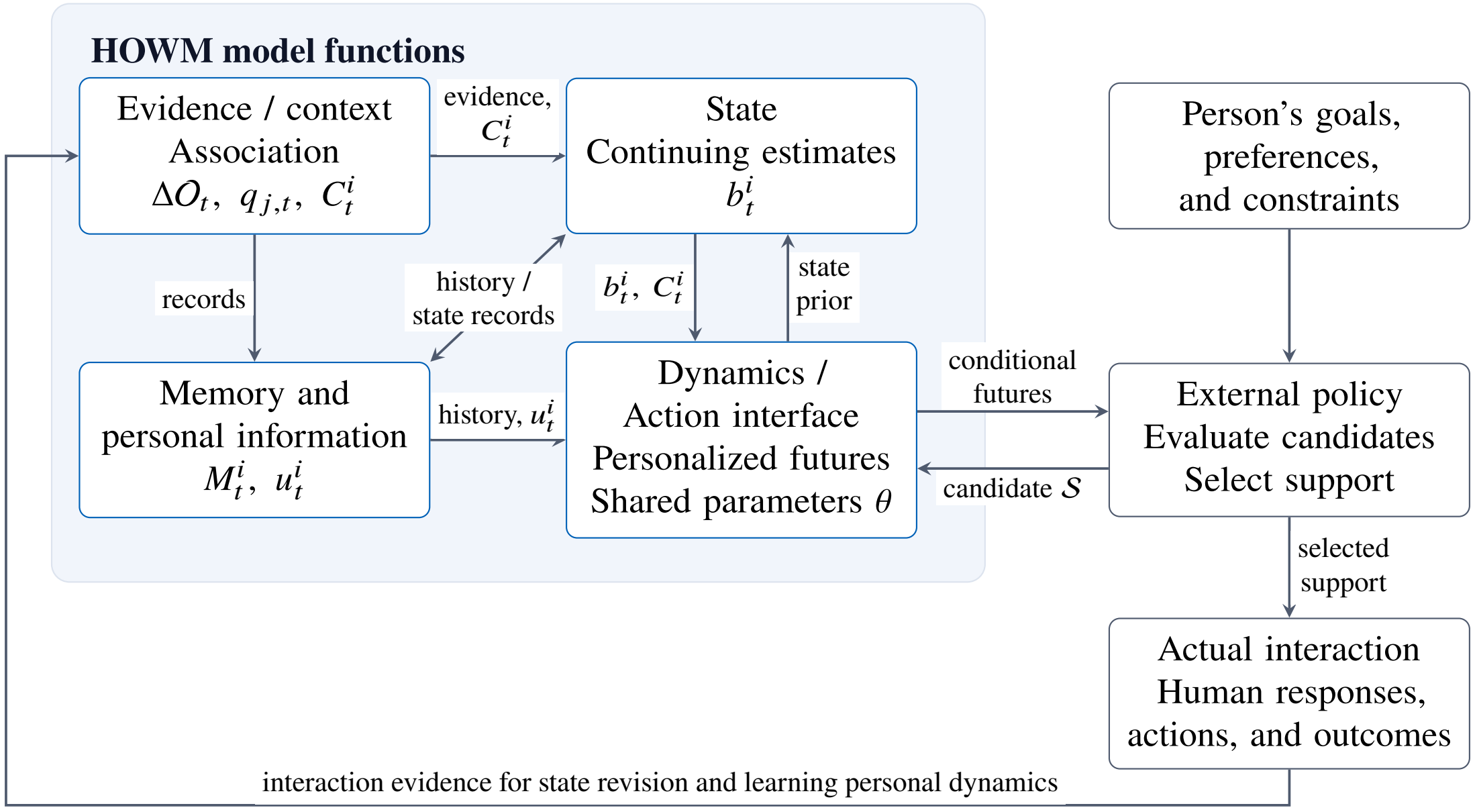
\begin{figure}[!htbp]
    \centering
    \begin{tikzpicture}[
        x=1cm,y=1cm,>=stealth,
        font=\small,
        component/.style={draw=finaiblue,fill=white,rounded corners=3pt,
            align=center,text width=3.1cm,minimum height=1.05cm,inner sep=5pt},
        external/.style={component,draw=finaisoft,text width=3.2cm},
        flow/.style={->,draw=finaisoft,line width=0.7pt},
        flowlabel/.style={font=\scriptsize,align=center,fill=white,inner sep=2pt}
    ]
        \draw[draw=finailine,fill=finaicard,rounded corners=5pt]
            (0,1.1) rectangle (9.2,6.8);
        \node[anchor=north west,font=\small\bfseries,text=finaiink]
            at (0.25,6.6) {HOWM model functions};
        \node[component] (evidence) at (2,5.3)
            {Evidence / context\\Association\\$\Delta\mathcal O_t,\ q_{j,t},\ C_t^i$};
        \node[component] (state) at (6.8,5.3)
            {State\\Continuing estimates\\$b_t^i$};
        \node[component] (memory) at (2,2.5)
            {Memory and personal information\\$M_t^i,\ u_t^i$};
        \node[component,minimum height=1.2cm] (dynamics) at (6.8,2.5)
            {Dynamics /\\Action interface\\Personalized futures\\Shared parameters $\theta$};
        \node[external] (goals) at (12.2,5.3)
            {Person's goals,\\preferences,\\and constraints};
        \node[external,minimum height=1.2cm] (policy) at (12.2,2.5)
            {External policy\\Evaluate candidates\\Select support};
        \node[external] (interaction) at (12.2,0)
            {Actual interaction\\Human responses, actions, and outcomes};
        \draw[flow] (evidence.east) -- node[flowlabel,above]
            {evidence,\\$C_t^i$} (state.west);
        \draw[flow] (evidence.south) -- node[flowlabel,left]
            {records} (memory.north);
        \draw[flow,<->] (memory.north east) -- node[flowlabel,xshift=-4pt]
            {history /\\state records} (state.south west);
        \draw[flow] (memory.east) -- node[flowlabel,above]
            {history, $u_t^i$} (dynamics.west);
        \draw[flow] ([xshift=-13pt]state.south) -- node[flowlabel,left]
            {$b_t^i,\ C_t^i$} ([xshift=-13pt]dynamics.north);
        \draw[flow] ([xshift=13pt]dynamics.north) -- node[flowlabel,right]
            {state\\prior} ([xshift=13pt]state.south);
        \draw[flow] ([yshift=8pt]dynamics.east) -- node[flowlabel,above]
            {conditional\\futures} ([yshift=8pt]policy.west);
        \draw[flow] ([yshift=-8pt]policy.west) -- node[flowlabel,below]
            {candidate $\mathcal S$} ([yshift=-8pt]dynamics.east);
        \draw[flow] (goals.south) -- (policy.north);
        \draw[flow] (policy.south) -- node[flowlabel,right]
            {selected\\support} (interaction.north);
        \draw[flow] (interaction.south) -- (12.2,-1.1) -- (-0.45,-1.1)
            -- (-0.45,5.3) -- (evidence.west);
        \node[flowlabel,above] at (6,-1.1)
            {interaction evidence for state revision and learning personal dynamics};
    \end{tikzpicture}
    \caption{Functional organization of HOWM. State and Dynamics connect continuing estimates to personalized futures using history and context. An external policy relates candidate futures to the person's goals and selects support. Actual interaction returns evidence for state revision and learning personal dynamics, which can inform subsequent estimation, prediction, and support. Arrows indicate information dependencies and allow shared or iterative computation.}
    \label{fig:howm_architecture}
\end{figure}

Figure~\ref{fig:howm_architecture} organizes the three capabilities through four functional responsibilities.
\emph{State} maintains current estimates; \emph{Memory} retains and retrieves individual experience; \emph{Dynamics} supplies expectations about change; and the \emph{Action} interface connects conditional forecasts to an external policy.
Dynamics advances an earlier estimate, new observations revise it, and the resulting estimate anchors predictions of what may happen next.
The policy relates these predictions to the person's goals, preferences, and constraints when selecting support.
Actual responses and outcomes provide evidence for revising the current estimate and, over repeated encounters, learning personal patterns of change.
Memory connects each encounter to applicable experience; these responsibilities allow shared representations and iterative computation.

An instance specifies its state variables, observation sources, historical information, prediction targets, and interaction setting.
It may use explicit variables, latent representations, or both, with one modality or several and with identity supplied by registration or inferred from observations.
The functional organization permits these implementation choices while keeping the meaning and use of the information explicit.
The following subsections develop the objects, mechanisms, and modeling choices involved.

\subsection{Modeling Objects and Information Roles}
\label{sec:framework_objects}

For ease of reference, we summarize HOWM's core notation and information roles in Table~\ref{tab:howm_notation}.

\begin{table}[tbp]
    \centering
    \caption{Core notation and information roles in HOWM.}
    \label{tab:howm_notation}
    \small
    \renewcommand{\arraystretch}{1.15}
    \begin{tabularx}{\linewidth}{@{}>{\raggedright\arraybackslash}p{0.22\linewidth}>{\raggedright\arraybackslash}X@{}}
        \toprule
        Symbol & Meaning and role \\
        \midrule
        $\mathcal I_t,\ q_{j,t}$ & Identifiers and uncertain assignments of observations or references to individuals. \\
        $Z_t=(H_t,E_t,R_t)$ & Task-relevant human, environmental, and relational states being modeled. \\
        $b_t,\ b_t^i$ & System estimates of the joint and individual states, including uncertainty. \\
        $\mathcal O_{\le t},\ \Delta\mathcal O_t$ & Evidence available by time $t$ and newly available evidence at the current update. \\
        $C_t^i$ & Available information or estimates about the person's environmental and social context. \\
        $M_t^i$ & Retained evidence and summaries with sources, times, and applicability conditions. \\
        $u_t^i$ & Personal characteristics, baselines, and learned patterns or adaptation information formed from history. \\
        $\mathcal P_i(t)$ & Individual model comprising $b_t^i$, $M_t^i$, and $u_t^i$. \\
        $\theta$ & Shared dynamics parameters used across individuals. \\
        $\mathcal S_{t:t+\tau}$ & Candidate future scenario specifying action or policy conditions and other assumptions. \\
        $Y_{t+1:t+\tau}^i$ & Future observable outcomes at the selected temporal resolution and horizon. \\
        \bottomrule
    \end{tabularx}
\end{table}

Let $\mathcal I_t$ denote the maintained individual identifiers, and write the task-relevant state as $Z_t=(H_t,E_t,R_t)$.
Here, $H_t=\{H_t^i\}_{i\in\mathcal I_t}$ contains human states, $E_t$ represents environmental and task conditions, and $R_t$ represents relevant relations and participant roles.
The system maintains an approximate estimate $b_t$ of $Z_t$, with $b_t^i$ denoting the corresponding individual estimate.
These estimates carry uncertainty about the modeled quantities; point or latent representations need a specified way to express that uncertainty.
A person's own beliefs may be components of $H_t^i$, whereas $b_t^i$ describes the system's estimate of the person's state.

The available observations and records, $\mathcal O_{\le t}$, include signals, actions that occurred, reports, and corrections received by time $t$.
Their sources, event times, and information-availability times distinguish what happened from when the system could use evidence about it.
Available context $C_t^i$ carries information or estimates about the environmental and social conditions relevant to person $i$, including those represented by $E_t$ and $R_t$.
This context helps interpret current behavior, conditions expectations about change, and constrains feasible support.
For example, a partner's cue can explain a pause, while the partner's pacing also affects subsequent demands and the arrangements available for the next exercise.
Observations, contextual interpretations, and state estimates can share evidence and associated uncertainty.

Individual memory $M_t^i$ retains historical evidence, summaries, and temporal relationships, together with their sources and conditions of applicability.
Personal information $u_t^i$ denotes characteristics, baselines, or adaptation information formed from this history, including representations of recurring patterns of change.
For example, typical performance provides a personal reference, while sequences around exercise and rest can inform expectations about how performance develops.
The current estimate $b_t^i$ describes the person's condition at a particular time; $M_t^i$ and $u_t^i$ connect that estimate to accumulated experience.
Retained reports and model interpretations preserve their evidential status so that these connections can be revisited.

These objects form an individual model,
\begin{equation}
    \mathcal P_i(t)=\bigl(b_t^i,M_t^i,u_t^i\bigr).
    \label{eq:individual_model}
\end{equation}
Shared dynamics parameters $\theta$ represent knowledge used across individuals for state propagation and future prediction.
Personalization can condition that shared model on $b_t^i$, experience in $M_t^i$, and learned information in $u_t^i$, including individual adaptation components where used.
The components of $\mathcal P_i(t)$ and the context may depend on the same records; their distinct symbols identify information roles without requiring separate storage or statistical independence.

Individual association connects these records to the corresponding person.
Reliable registration or confirmation can establish a link; otherwise, $q_{j,t}$ denotes an approximate assignment distribution for observation or reference $j$ over $\mathcal I_t\cup\{\mathrm{new}\}$, including a person not yet represented by a maintained identifier.
Identifiers and uncertain links can be revised as evidence arrives, with roles retained for events involving several people.
Keeping the basis of an association available helps prevent agreement with an existing personal model from becoming its own supporting evidence.

The observation window, retained historical span, and future prediction horizon are separate temporal choices.
Observations may be irregular or interrupted, and a long history may inform a short forecast.
New information about an earlier event can revise its interpretation while preserving what was available when an earlier estimate or forecast was produced.
This distinction allows the model to follow changes in the person and to revise its understanding of the same past episode.

\subsection{Continuing State Estimation in Context}
\label{sec:framework_estimation}

Continuing estimation follows the person through intervals with and without new observations.
Between observations, dynamics propagates the previous estimate using elapsed time, personal patterns, and known activities or interactions.
After a rest period, for example, the expectation depends on the earlier condition, the duration of rest, and relevant personal experience.
New movement records, reports, and contextual information then revise that expectation to form $b_t^i$, which becomes the starting point for subsequent forecasts.
Unknown intervening events contribute uncertainty about the state reached, so an observation gap also affects how confidently the model can describe the present.

Personal history and context help distinguish the explanations compatible with a new observation.
A baseline locates performance relative to comparable earlier episodes, while the recent trajectory indicates whether a deviation is emerging, persisting, or reversing.
An apparent hesitation during a partner-paced exercise may reflect waiting for a cue, and a change of task can alter performance even when the underlying personal condition is similar.
The estimate therefore combines what was expected to change with evidence about the task and interaction actually taking place.
If several explanations remain plausible, their uncertainty carries forward into predictions and support decisions.

Maintaining this process requires representations that preserve changes relevant to the task and the relationship between each observation and the modeled state.
Movement records and reports may concern different aspects or times, while multiple interpretations of one recording share an evidential basis.
Reliable association and contextual applicability determine which historical evidence can inform the estimate.
When new observations repeatedly conflict with expectations, the model may need to revise the current state estimate, its interpretation of the context, or the personal pattern used for propagation.
The revised estimate then provides a new basis for following the person's subsequent development.

\subsection{Personalized Dynamics Prediction}
\label{sec:framework_prediction}

Individual dynamics describes possible changes from a person's current state under specified conditions.
The target may be a state trajectory, an observable outcome $Y_t^i$, or a sequence of events at a chosen temporal resolution.
An observed result such as exercise completion can reflect the person's condition together with task demands without revealing every component of that condition.
Let $\mathcal S_{t:t+\tau}$ denote a candidate scenario over horizon $\tau$, specifying action or policy conditions and relevant assumptions about the environment and interaction.
Here, $t+1$ denotes the next point on the chosen prediction grid, with $\tau$ counting future steps.
Actual elapsed times are specified by timestamps; observations need not arrive at every grid point.
When both states and observable outcomes are represented, the prediction interface is
\begin{equation}
    p_\theta\!\left(
    H_{t+1:t+\tau}^i,Y_{t+1:t+\tau}^i
    \mid \mathcal P_i(t),C_t^i,\mathcal S_{t:t+\tau}
    \right).
    \label{eq:conditional_future}
\end{equation}
An instance may instead predict only one type of target, using the corresponding marginal when a joint model is available.
The scenario distinguishes conditions supplied to the forecast from responses and changes that remain to be predicted or marginalized.
Uncertainty in the starting estimate and context therefore remains relevant alongside uncertainty about the future.

Personalized dynamics connects a starting condition to a possible trajectory for this individual.
A baseline describes a typical level, a current deviation locates the present relative to that level, and a dynamic pattern describes how the state or performance may subsequently change.
Two people with similar current movement stability may have histories suggesting different rates of decline during continued exercise or recovery after rest.
For the same person, a brief deviation following a task switch may suggest a different development from a gradual decline across several exercises.
Such distinctions concern the direction, rate, persistence, and uncertainty of change under specified conditions, and determine which parts of an individual history matter to the forecast.

Environmental and social context helps determine the trajectory being considered.
Exercise demands, rest opportunities, a partner's pace, and instructions can affect the development associated with the same starting condition.
Prediction therefore uses current context together with the assumed future arrangement in $\mathcal S_{t:t+\tau}$; changes in another participant's behavior may themselves remain part of the uncertain future.
For example, a forecast under continuing coordination with a partner and one under independent pacing concern different conditions.
Historical episodes inform these forecasts to the extent that their starting states, task demands, and interactions are applicable to the current question.

Shared parameters provide an initial basis when individual experience is sparse.
As relevant trajectories accumulate, the model can condition predictions on comparable episodes or adapt the personal information in $u_t^i$ to reflect recurring patterns.
For example, repeated observations around documented rest periods may support a more specific expectation about this person's subsequent performance than the shared model alone.
Recent development may already be represented in $b_t^i$ and recurring patterns in $u_t^i$; retrieval from $M_t^i$ supplies further detail when those summaries omit information needed for the target.
Retaining event order, elapsed time, and contextual conditions makes it possible to connect observed changes to their circumstances and reconsider a pattern when later evidence differs.

The same process can connect several time scales.
Movement within an exercise, accumulation and recovery during a session, and development across sessions concern different trajectories and prediction targets.
Longer-term experience can condition a short forecast, while repeated short trajectories can reveal changes in the person's baseline or response pattern across sessions.
A session average preserves a typical level but may omit the order of fluctuations needed to predict the next exercise; aggregation and memory selection should therefore preserve the temporal relationships required by the target.

\subsection{Conditional Prediction and Action Support}
\label{sec:framework_action}

Action support starts from the person's goals and the needs suggested by the current situation.
The relevant futures depend on what the person wishes to achieve, their preferences, and practical constraints.
For the same predicted training trajectories, completing a current exercise and maintaining comfortable participation over a longer session may place different emphasis on immediate performance, effort, and later readiness.
An inferred goal can help explain behavior, while the person's expressed priorities provide grounds for evaluating proposed support and can change during the interaction.

Through the Action interface, an external policy supplies candidate scenarios and receives forecasts of relevant developments with their assumptions and uncertainty.
Continuing an exercise, proposing a slower set, or taking a break can involve different expected trajectories of performance and effort.
Environmental and social context determines which arrangements are feasible, such as whether changing pace requires coordination with a partner.
For a proposal, acceptance, modification, or rejection may be part of the predicted future; a forecast conditional on acceptance instead treats that response as a scenario assumption.
The policy relates supported predictions under these conditions to the person's priorities when considering what to offer.

Current estimates and forecasts may suggest an emerging difficulty or an opportunity to support the person's expressed goals, even without a new explicit request.
Within an agreed scope of assistance, the external policy can use this evidence to initiate an offer of support.
For example, a forecast of increasing effort during continued exercise may prompt the policy to offer an adjustment before the next set.
The policy considers the person's preferences and the burden of interruption when deciding how and when to make the offer.

Interaction can help clarify uncertainty relevant to the choice of support.
If alternative interpretations of a pause would lead to different suggestions, asking about difficulty or the expected cue may help clarify the situation.
The policy can consider the expected value of that information together with the burden of asking, or wait for further evidence when appropriate.
Such interactions connect assistance to learning: the response can inform both the immediate choice and the model's interpretation of similar situations later.

Actual participation determines which arrangement occurs and what can subsequently be learned from it.
The interaction record distinguishes the proposed support, the person's response, the action taken, and immediate or delayed outcomes.
A modified exercise is interpreted under the conditions actually delivered, and predicted acceptance remains distinct from the person's agreement.
Selection and execution belong to the policy and the actual interaction, while the model supplies information about possible developments.
Section~\ref{sec:research_assistance} specifies how forecast quality and assistance outcomes are assessed, including the additional evidence needed to interpret alternative actions as interventions or counterfactuals.

\subsection{Interaction Feedback and Continuing Model Updates}
\label{sec:framework_updates}

Actual interaction connects an anticipated future to what the person subsequently does and experiences.
Reports, observed responses, and outcomes supply evidence for the next state estimate and for examining expectations formed earlier.
Asking about difficulty can clarify the present condition, while completing or modifying an exercise supplies a trajectory under the conditions that actually occurred.
Feedback is interpreted with those conditions and the observation process: more frequent questions, for example, can increase reports without a corresponding change in the underlying state.
Across encounters, these records provide the experience from which the model can refine its understanding of the person.

The update interface connects this experience to the maintained individual models,
\begin{equation}
    \{\mathcal P_i(t)\}_{i\in\mathcal I_t}
    =\mathcal U_\theta\!\left(
    \{\mathcal P_i(t^-)\}_{i\in\mathcal I_{t^-}},
    \Delta\mathcal O_t,\{q_{j,t}\}_j,\{C_t^i\}_{i\in\mathcal I_t}
    \right),
    \label{eq:model_maintenance}
\end{equation}
where $t^-$ is the system time before the update and $\Delta\mathcal O_t$ contains newly available observations, feedback, or corrections.
The retained estimates carry their reference times, allowing elapsed time to enter state propagation using the current shared and personal information.
Association and context can themselves be inferred using the individual models, so these dependencies may be resolved jointly or iteratively, including revisions to identifiers and histories.

The contents updated depend on what the new experience establishes.
A report about present effort can revise $b_t^i$, and the report with its interaction context extends $M_t^i$.
Comparable trajectories can inform baselines or personal dynamic patterns represented in $u_t^i$, while a corresponding learning process can revise shared parameters $\theta$ across individuals.
Changes to support selection belong to the external policy.
A single encounter may refine only the current estimate; learning a recurring pattern requires evidence across relevant starting, task, and action conditions.

For example, suppose performance repeatedly returns toward the person's usual level sooner after rest than the model anticipated under comparable conditions.
Those observations may support revising the personal recovery pattern, after considering changes in task demands and measurement conditions.
On a later encounter, the revised pattern can alter the state prior after a rest interval, the forecast for the next exercise, and the information supplied when considering further support.
If later trajectories depart from that pattern, the model can revisit whether the person has changed, the contextual conditions differ, or its earlier understanding was incomplete.
This connects accumulated feedback to subsequent use without requiring an explicit correction.

Corrections propagate through the contents that depended on the revised evidence.
A later account of an instruction may change the interpretation of an earlier observation and the estimates or summaries formed from it.
A corrected individual assignment may affect several personal histories and shared events.
The scope of revision depends on these relationships and the evidence still recoverable.
Original forecasts remain linked to their original information basis, while later interpretations carry their own availability times.

Continuing updates thus connect current estimation, personal dynamics, and later support through the same evolving body of experience.
Whether these updates improve understanding is an empirical question about subsequent estimates and forecasts; whether their use helps the person additionally depends on the support chosen and its outcomes.
The next section turns the mechanisms into research questions and comparison designs for assessing estimates, forecasts, updates, and the outcomes of assistance.

\section{Research Questions and Validation Pathways}
\label{sec:research_agenda}

\subsection{Research Questions and Assessment Targets}
\label{sec:research_targets}

The framework raises three capability questions.
Continuing estimation asks whether the model follows a person's changing state and revises its understanding as evidence arrives.
Personalized prediction asks whether it anticipates that individual's development under specified task and interaction conditions.
Action support asks whether using those forecasts helps the person pursue their goals.
Learning from actual interaction runs across all three: accumulated experience should be assessed through its effect on subsequent estimates, predictions, and support.

A study first specifies the task, modeled state variables, available observations and context, prediction targets, and assessment evidence.
For state estimation, an independent, task-appropriate reference can support claims about accuracy for the variables and times it covers.
Where only observable consequences are available, the study instead evaluates how the estimate supports predictions of those consequences.
Any direct contribution of history to the prediction needs to be distinguished from its contribution through the estimate.
A further claim about state-estimation accuracy requires a justified measurement relation or independent validation of the estimate.
Reference quality and uncertainty therefore belong in the task specification.

Assessment follows these capabilities through time.
For estimation, relevant targets include successive errors, calibration, and the time or evidence needed to recover from an inaccurate estimate, including after observation gaps or task changes.
For prediction, studies specify the trajectory or observable outcome, horizon, and temporal resolution, distinguishing absolute level, deviation from a personal baseline, and subsequent change.
For support, they follow the person's choices and outcomes in relation to their goals; for updating, they examine how later use changes as experience accumulates.
Table~\ref{tab:research_questions} connects these questions to comparison designs.
H1--H3 investigate particular historical and action-information contributions within these studies.

\subsection{Testable Hypotheses and Comparative Designs}
\label{sec:research_comparisons}

Capability studies first identify the process whose contribution is being examined.
Continuing estimation can compare models with and without state propagation or contextual information under common observation sequences, including gaps and changes of task or social arrangement.
Personalized prediction can compare a shared predictor, one with a personal baseline, one using individual temporal patterns, and one that adapts those patterns as new experience arrives.
These comparisons ask whether the model can track and anticipate change, while H1--H3 below examine how specific information sources contribute.
Support outcomes require separate policy comparisons.

Each comparison declares the information available to its conditions, including history already encoded in estimates, baselines, parameters, or other summaries.
Model capacity, training access, current evidence, and relevant contextual coverage should be comparable except for the contribution under study.
Association reliability and historical applicability need operational criteria, such as confirmed identity and recorded task conditions, established independently of the eventual estimation or prediction scores.
These controls allow gains and failures to be related to a specified modeling choice and the conditions in which it is used.

\textbf{H1: Personal history in state estimation.}
Personal history available at the estimation time can improve current-state estimation when current observations are ambiguous and the history supplies applicable individual evidence.
A comparison uses current observations and context in both conditions and adds specified personal history in one.
If both conditions already receive a personal baseline, the comparison concerns the additional value of retained records; a study of the total contribution of history must also account for its effect through that baseline and other summaries.
The assessment follows the reference requirements in Section~\ref{sec:research_targets}, and can examine both successive estimates and their uncertainty.
Current observations may suffice, while outdated or incorrectly associated history may reduce accuracy.

\textbf{H2: Individual history and personalized dynamics.}
Applicable personal history can improve future prediction beyond shared knowledge and current observations, with temporal patterns potentially contributing beyond a personal baseline.
A baseline comparison examines the value of knowing the person's typical level; a dynamics comparison examines the further value of individual progression, recovery, or responses under specified conditions.
Targets can include the direction and extent of later change, a trajectory across an activity, or an outcome following a task transition at a common horizon.
When the intended claim concerns dynamics beyond knowledge of the starting state, the predictors receive a common current estimate and baseline.
A study of the total contribution of history also allows it to affect the starting estimate and declares those pathways.

Historical span and temporal structure can then be examined as contributions to these predictions.
One comparison varies access to recent or longer-term experience at the same target and horizon, accounting for record quantity and contextual coverage.
Another varies access to event order, elapsed time, or organization across sessions while retaining comparable record content and context.
Their contribution can differ across within-session and across-session targets, or when task demands and social arrangements change.
These comparisons connect retained experience to the temporal relationships used for prediction.
An improvement supports the stated forecasting capability, while identifying an underlying transition mechanism requires evidence that distinguishes alternative explanations.

\textbf{H3: Action information in conditional prediction.}
Explicit action conditions can improve predictions of subsequent human states, responses, and outcomes when those conditions are relevant to the target, with applicable personal response history potentially providing further value.
The principal comparison includes or omits the specified action information under comparable current information and modeling conditions; an additional comparison varies access to personal response history.
Each condition must identify action information already present in task arrangements, context, or other representations, since an explicit action input may otherwise be redundant.
Assessment uses realized responses and outcomes under conditions corresponding to the forecast.
The evidence required for comparing unchosen alternatives and for evaluating actual support is developed in Section~\ref{sec:research_assistance}.

\begin{table}[tbp]
    \centering
    \caption{Capability questions and corresponding validation designs. H1--H3 investigate information contributions within this agenda; continued updating is assessed through later use across the three capabilities.}
    \label{tab:research_questions}
    \small
    \renewcommand{\arraystretch}{1.15}
    \begin{tabularx}{\linewidth}{@{}>{\raggedright\arraybackslash}p{0.17\linewidth}>{\raggedright\arraybackslash}X>{\raggedright\arraybackslash}X@{}}
        \toprule
        Focus & Question and comparison & Assessment targets \\
        \midrule
        Continuing estimation & Does the estimate follow change? Vary state propagation, context, or personal history (H1) under common observations. & Successive estimates, uncertainty, and recovery after gaps or task changes, using appropriate state references. \\
        \addlinespace
        Personalized dynamics & Does the model anticipate individual development? Compare shared, personal-baseline, and temporal-pattern predictors; examine historical contributions~(H2). & Future change and trajectories at common origins and horizons, with starting-state information and contextual coverage declared. \\
        \addlinespace
        Action support & Does using forecasts help the person? Compare support policies and separately examine action-conditioned prediction (H3). & Forecast quality under realized conditions; choices, burden, and outcomes relative to the person's goals. \\
        \addlinespace
        Continued updating & Does experience improve later use? Compare adapted and retained personal patterns on common later observations; assess corrections. & Later estimation and prediction, adaptation to change, recurring errors, and usefulness of subsequent support. \\
        \bottomrule
    \end{tabularx}
\end{table}

\subsection{Longitudinal Evaluation of Model Updating}
\label{sec:research_updates}

A longitudinal design follows the same individuals through repeated observations, interactions, and model updates.
One comparison allows a personal dynamic pattern to adapt while another retains its earlier version; both continue to receive the same current observations and context at later prediction origins.
Their current estimates may evolve accordingly when the study assesses the total downstream effect of learning.
A narrower prediction comparison supplies a common starting estimate to isolate the additional contribution of the updated dynamic pattern.
Assessment tracks later estimates, forecast quality, and uncertainty as relevant experience accumulates, separating the benefit of learning from merely receiving a new observation.

Progress is examined at subsequent encounters with comparable conditions as well as after documented changes.
For example, observations around rest periods can test whether learning a personal recovery pattern improves later state priors and forecasts for the next exercise.
After a correction to an instruction or contextual record, a study can assess both the dependent historical revisions and whether similar misinterpretations recur in later encounters.
Time or experience needed to adapt, persistent errors, and uncertainty under unfamiliar conditions provide complementary targets.
When the updated model is used for support, its effects on later choices and human outcomes require separate evaluation.

Every estimation or prediction origin defines an information boundary for its inputs, summaries, adaptations, and shared parameters.
Reconstructing an original forecast requires the information and model version available then; later assessment evidence may supply a reference without becoming an earlier input.
Retrospective estimates using a correction are evaluated separately, including whether dependent contents were revised appropriately and unrelated contents remain consistent.
An evaluation of a model trained later declares its training scope, data partitions, and temporal restrictions as a separate protocol.

Learning curves need interpretation alongside changes in task difficulty, contextual coverage, and how evidence becomes available.
Changed devices, missing periods, different partners, and more frequent questioning can affect the observed record and the comparability of later assessments.
Matched prediction origins and documented conditions help distinguish learning progress from an easier evaluation period or a changed observation process.
If the evidence cannot distinguish personal change from altered context or measurement, that ambiguity remains part of the result.

\subsection{Action Support and Human Outcomes}
\label{sec:research_assistance}

Evaluating assistance follows the support offered, the person's response, the action actually taken, and the immediate or delayed outcomes relevant to their goals.
The study specifies those goals and an observation period appropriate to the claimed benefit, recording changes in priorities where they affect support choices.
Task outcomes, experienced burden, willingness to continue, and the person's assessment of the support can provide complementary evidence.
For proactive support, evaluation also examines the relevance and timing of offers made without a new request and the burden of interruption.
Acceptance records participation in a proposal; the subsequent outcomes establish what happened under the arrangement that was actually chosen.

Where an appropriate design is available, support selected using forecasts can be compared with a specified alternative policy under comparable opportunities and resources.
The study examines whether predictions change the timing or form of support and whether the resulting choices help the person pursue the stated goals.
For example, it can distinguish improved forecasts that leave choices unchanged from changes in support that are followed by different outcomes.
A longitudinal comparison can further examine whether incorporating new interaction experience improves later support; the scope of model and policy updates is declared for each condition.
Because support may also change how feedback is obtained, outcomes are interpreted alongside the observation process.

Forecast quality receives its own assessment against the conditions specified at prediction time and the responses or outcomes subsequently observed.
A forecast of acceptance leaves the person's choice unresolved; a forecast conditional on acceptance treats that choice as given.
The later choice cannot be supplied as known information when the task is to predict it.
Linking forecasts, choices, and outcomes makes it possible to examine where a useful prediction enters the assistance process while keeping their respective evaluation targets explicit.

The required evidence depends on whether a conclusion concerns conditional prediction, an intervention effect, or an individual counterfactual.
Observed action--outcome associations can support forecasting under the studied conditions, while claims about changing the action require causal assumptions and identification evidence.
Historical support may have been selected in response to unrecorded aspects of the person's condition, and unchosen arrangements need evidence beyond the outcome of the selected action.
The design therefore specifies the action conditions and population or individuals covered by the comparison, together with the assumptions supporting any causal interpretation.

\subsection{Data Requirements and Staged Validation}
\label{sec:research_validation}

Longitudinal records need to link observations and events to individuals, task and environmental conditions, participant roles, and the references or later outcomes relevant to each claim.
Event time and information-availability time establish which records could be used at a prediction origin.
Action studies additionally distinguish proposed support, the person's choice, the action actually taken, and immediate or delayed outcomes.
Records also distinguish offers made without a new explicit request from those made in response to one.
Association uncertainty, observation conditions, and corrections remain attached to the corresponding evidence.
These records support the information controls and version requirements described above.

An initial study can select a bounded task and state subset with suitable observations and assessment evidence.
It runs a continuing sequence of observation and context, state estimation, a limited future prediction, subsequent evidence, and updating before the next encounter.
A lightweight predictor of a specified later state or outcome provides an initial dynamics component, compared with a suitable simple baseline at successive prediction origins.
When it predicts a modeled state, its forecast can also supply a prior for the next estimation step; predicting only an observable outcome supports assessment of that outcome forecast.
Activities, reports, and responses are recorded from the beginning so that predictions and updates remain connected to the conditions that actually occurred.
This initial process establishes a setting for H1 and H2 comparisons and for identifying limitations in the target, representation, history, learning method, or data coverage.

As trajectories accumulate, studies can develop personal dynamic patterns, assess adaptation, and expand action-conditioned prediction under documented interaction arrangements.
Using the forecasts to consider support then enables studies of choices and human outcomes.
Table~\ref{tab:validation_pathways} relates these studies to their data and assessment requirements.

\begin{table}[!htbp]
    \centering
    \caption{Validation pathways and required data. Stages may overlap; updating is assessed in each capability.}
    \label{tab:validation_pathways}
    \small
    \renewcommand{\arraystretch}{1.15}
    \begin{tabularx}{\linewidth}{@{}>{\raggedright\arraybackslash}p{0.18\linewidth}>{\raggedright\arraybackslash}X>{\raggedright\arraybackslash}X@{}}
        \toprule
        Study focus & Required evidence & Validation targets \\
        \midrule
        Continuing estimation and initial prediction & Repeated observations, context, activities, times, state references or later outcomes. & Successive estimates and forecasts followed by new evidence and updating; simple baselines and H1 comparisons. \\
        \addlinespace
        Personalized dynamics & Individual trajectories across activities or sessions, relevant conditions, and outcomes. & Prediction of change beyond personal baselines; H2 historical contributions and adaptation with experience. \\
        \addlinespace
        Conditional prediction and assistance & Proposals, human choices, executed actions, outcomes, goals, and observation conditions. & H3 predictive contribution; separate policy comparisons and human outcome assessment. \\
        \addlinespace
        Learning and revision across stages & Successive model and information versions, interaction feedback, corrections, later observations, and references. & Later estimation, prediction, and support; adaptation and recurring errors, with original forecasts preserved. \\
        \bottomrule
    \end{tabularx}
\end{table}

The stages express research dependencies and can overlap, with subsequent observations and interaction feedback supporting updates throughout.
A primary longitudinal setting can support sustained investigation, while supplementary settings examine the particular functions and conditions they instantiate.
The next section illustrates this organization in hypothetical training and learning settings.

\section{Illustrative Scenarios}
\label{sec:scenarios}

\subsection{Sports and Training: A Continuous Example}
\label{sec:scenarios_training}

Consider a hypothetical badminton participant whose expressed aim is to practice consistent footwork at a manageable level of effort.
The model follows movement control and the participant's current training condition, using movement timing, stability, and reports of effort as evidence.
It considers how those states and the resulting performance may develop through the next exercise and how that development relates to the participant's aim.
A coach sets the exercise and cue schedule, supplying environmental and social conditions that affect both the interpretation of movement and the available practice arrangements.
The example illustrates the framework's use.

At $t_0$, reliably associated earlier sessions provide a personal history $M_{t_0}^i$ with exercise conditions, cue schedules, reports, event times, and information-availability times.
Suppose comparable sequences show that this participant often hesitates after a change in cue pace, with timing becoming more regular over subsequent repetitions.
Sequences under a familiar slower schedule suggest a shorter period of adjustment under otherwise comparable conditions.
These hypothetical records support both a baseline for typical performance and an initial personal pattern of change in $u_{t_0}^i$.
The baseline locates a current deviation; the pattern helps anticipate how that deviation may develop.
The pattern's applicability to a new session is uncertain.

Before the opening observations at $t_1$, the model propagates its latest estimate using elapsed time and known intervening activity, together with shared and personal dynamic information.
Unknown activity during the gap contributes uncertainty.
Suppose the opening exercise introduces a faster cue pace.
The participant shows hesitations and lower stability than the relevant baseline, while reporting greater difficulty than usual.
The cue arrangement in $C_{t_1}^i$ and earlier adjustment sequences help interpret these observations when forming $b_{t_1}^i$.
The estimate allows for adjustment to the new timing as well as a change in the participant's training condition; their relative contributions remain uncertain.
This estimate and the observed progression within the exercise anchor the next forecast.

For continuation at the current cue pace, suppose the model anticipates that hesitations may persist early in the next set and then diminish as the participant adjusts.
For a return to the familiar slower schedule, it anticipates more regular timing earlier in the set, while uncertainty about the current condition still affects predicted stability and completion.
These forecasts use personal history to describe the direction and timing of change under each arrangement.
For this illustration, both arrangements fall within the conditions covered by earlier records and prediction assessments.
The model retains the forecasts with their information basis and assumptions; outcomes conditional on an arrangement being accepted remain separate from any prediction of the participant's acceptance, modification, or refusal.

The policy considers these forecasts together with the participant's aim of consistent movement at manageable effort.
Suppose the coach confirms that the slower schedule is feasible.
The prospect of steadier timing makes it a candidate for supporting the stated aim, while the participant's assessment of expected effort and willingness informs whether it should be tried.
Suppose the participant has agreed to occasional suggestions between sets.
At the next between-set pause, the policy offers that option without a new request and asks whether it feels manageable; the participant can accept, modify, or decline it.
If the participant's account instead makes the cue interpretation uncertain, clarification can precede a revised forecast and proposal.

At $t_2$, suppose the participant accepts the slower set, completes it with steadier movement, and reports manageable effort.
The record links the proposal and choice to the cue schedule actually delivered, the movement trajectory, and the participant's report.
The trajectory can be compared with the forecast under that arrangement, and the new evidence revises $b_{t_2}^i$ and extends $M_{t_2}^i$.
Steadier movement under reduced demands informs the estimate under those conditions; it leaves open how the participant would perform at the earlier pace and whether the broader training condition has improved.
This encounter contributes one contextualized trajectory to later learning, with any baseline revision depending on comparability to the exercises that baseline describes.

Across subsequent sessions, suppose comparable trajectories repeatedly stabilize earlier under the familiar schedule than the model anticipated, with task and observation conditions documented.
This experience can support revising the personal adjustment pattern, including its expected duration and uncertainty.
At a later encounter $t_3$, the revised pattern informs the state prior after the opening repetitions and the forecast for the remainder of the set; current observations and reports can confirm or challenge that expectation.
The policy can use the resulting forecast when deciding whether to ask about remaining difficulty or offer another feasible practice arrangement in relation to the participant's current aim.
Thus, ordinary responses and outcomes can alter later estimation, prediction, and support without requiring an explicit correction.

A supplementary revision branch arises if a later coach annotation shows that the opening cue timing at $t_1$ differed from its recorded plan.
The revised context prompts reconsideration of the earlier estimate and any personal pattern that incorporated that episode.
It may change how much of the hesitation is attributed to exercise demands, while leaving other explanations unresolved.
Original forecasts remain linked to their original information basis; the later interpretation informs subsequent use with its own availability time.

\subsection{Education and Learning: Progress across Activities}
\label{sec:scenarios_learning}

Consider a learner whose expressed goal is to solve a particular type of problem independently.
Reliably associated task responses, hints used, and learner feedback provide evidence about present understanding and experienced difficulty, with problem demands and teacher or peer assistance retained as context.
Suppose earlier sequences show successful completion soon after a demonstration but less consistent performance on later unprompted problems of comparable difficulty.
This history informs expectations about progress across activities as well as a baseline score.
New responses revise a starting estimate formed from the previous state estimate, elapsed time, and intervening learning, with uncertainty about knowledge gaps, temporary difficulty, or interpretation of the instructions.

From that estimate, the model forecasts performance over a specified sequence of activities, distinguishing tasks with hints from later independent attempts.
In this hypothetical case, it may anticipate success on the next supported problem while remaining uncertain about subsequent independent completion.
The policy relates these forecasts to the learner's goal and preferences when proposing another demonstration, a targeted hint, or an independent attempt with help available on request.
The learner's choice and actual use of assistance determine the conditions under which the next response is interpreted.
Success after a prompt supplies evidence about supported performance; later unprompted attempts help assess whether the understanding carries into the learner's intended setting.

Suppose the learner chooses a hint, completes the current problem, and later encounters difficulty on an independent attempt.
Those outcomes refine the current estimate and extend the history under their respective assistance conditions.
If comparable sequences subsequently show less need for hints and more consistent independent performance, the accumulated evidence can support revising the expected progression toward independent completion.
At the next activity, that revised pattern can change both the initial estimate and the forecast used to consider further support.
This example follows learning over a sequence of activities, with actual choices and outcomes continually revising the model's understanding of the learner's development.

\subsection{Common Structure and Conditions for Validation}
\label{sec:scenarios_validation}

The examples follow the same continuing process through task-specific states and patterns of change.
Training emphasizes adjustment within an exercise and the learning of personal response patterns across sessions; the learning example follows progress from supported to independent activity.
In both, observations and contextualized history inform the current estimate, personal dynamics connects that estimate to possible futures, and the person's goals give those futures their relevance for support.
Actual interactions then provide evidence that can change the next estimate, forecast, and support proposal.
Table~\ref{tab:scenario_mapping} summarizes these connections, with the state meanings, relevant conditions, and criteria for useful assistance specified within each domain.

\begin{table}[!htb]
    \centering
    \caption{State, personal dynamics, and continuing use in the hypothetical training and learning settings.}
    \label{tab:scenario_mapping}
    \small
    \renewcommand{\arraystretch}{1.15}
    \begin{tabularx}{\linewidth}{@{}>{\raggedright\arraybackslash}p{0.19\linewidth}>{\raggedright\arraybackslash}X>{\raggedright\arraybackslash}X@{}}
        \toprule
        Modeling role & Sports and training & Education and learning \\
        \midrule
        State and evidence & Movement control and training condition inferred from movement and reports. & Understanding and difficulty inferred from task responses and learner feedback. \\
        \addlinespace
        Individual history & Associated practice trajectories, cue schedules, reports, and personal baselines. & Linked task sequences, hints used, feedback, and performance on comparable material. \\
        \addlinespace
        Context & Exercise demands, court conditions, coach or partner roles, and cue pace. & Problem demands, instructions, and teacher or peer assistance. \\
        \addlinespace
        Personal dynamics & Timing and stability in adapting to cue schedules, with applicability and uncertainty. & Progress across supported and independent tasks, with applicability and uncertainty. \\
        \addlinespace
        Conditional futures & Movement trajectories and completion under the proposed practice arrangement. & Performance over a task sequence with specified assistance conditions. \\
        \addlinespace
        Goals and support & Consistent footwork at manageable effort informs pace changes or clarification. & Independent problem solving guides consideration of demonstrations, hints, or further attempts. \\
        \addlinespace
        Actual feedback & Choice, delivered cue schedule, movement trajectory, and experienced effort. & Choice, assistance actually used, task responses, and experienced difficulty. \\
        \addlinespace
        Subsequent use & New trajectories can revise the pattern used in later estimates, forecasts, and support. & New sequences can revise expected progress and forecasts informing later support. \\
        \bottomrule
    \end{tabularx}
\end{table}

These hypothetical sequences illustrate how HOWM could operate; they provide no empirical evidence of prediction gains or assistance benefits.
Turning them into studies requires longitudinal records and assessment evidence, with separate comparisons of later model use and the person's outcomes.

\section{Scope, Limitations, and Open Questions}
\label{sec:limitations}

\subsection{Task Scope and the Limits of State Inference}
\label{sec:limitations_state}

The framework organizes a task-relevant account of a person, whose adequacy depends on the variables selected and the situations covered.
Relevant aspects of the person's experience may remain outside that account, even when its predictions are useful within a task.
Changes in goals, relationships, or activities can make previously omitted quantities important and require reconsidering existing representations and summaries.
Extending an instance to a new setting therefore raises a substantive modeling question about which meanings and relationships remain applicable.
Environmental and social context is itself selective: unrecorded instructions, changing participant roles, or another person's unexpected behavior may alter the interpretation of an observation or the conditions of a forecast.

Within a chosen scope, some hidden states may remain indistinguishable from the available observations.
Different internal conditions can produce similar behavior, and personal reports and external observations may cover different aspects or times.
Predicting performance under a particular cue schedule or level of assistance can be useful while leaving uncertainty about the person's training condition or understanding.
Interpreting those states requires evidence about the relationship between observations and the modeled quantities.
The uncertainty expressed by a model may itself be inaccurate or omit plausible explanations.
When dynamic expectations inform continuing estimation, an inaccurate prior can persist if new evidence is weak or given insufficient weight.

\subsection{Personal History, Dynamics, and Continuing Learning}
\label{sec:limitations_history}

The available history limits which individual patterns can be learned and where they can be applied.
Many records from a narrow range of tasks or support conditions may provide little evidence about development in other circumstances.
With sparse individual experience, shared knowledge supplies much of the predictive basis, and estimates of personal differences can remain uncertain.
Incorrect associations, missing events, and unsupported summaries can further distort that basis and reinforce earlier interpretations.
Revision depends on whether the original evidence and contextual details remain recoverable, even when the dependencies between records are known.

Personal dynamic patterns can themselves change as the person learns, practices, encounters new circumstances, or revises their goals.
A baseline or adjustment pattern learned earlier may consequently become a poor guide to subsequent development.
Adapting quickly can incorporate transient fluctuations into a personal pattern, while adapting slowly can preserve an outdated expectation.
The appropriate balance between retained experience, shared knowledge, and recent evidence depends on the changes being modeled and the observations available to distinguish them.
Long histories therefore require attention to the continuing applicability of their contents as well as their span.

The observation process creates further gaps even when retained records are accurate.
Device changes, intermittent sampling, and altered task demands can produce apparent trends or obscure actual changes.
Records confined to activities the person chooses to complete also leave uncertainty about other situations.
Learning from these histories requires distinguishing changes in the person, changes in context or measurement, and revisions to the model's knowledge.
A contextual correction can improve an estimate of an earlier episode without establishing that the person changed at that time, and available evidence may leave these explanations partly unresolved.

\subsection{Actions, Human Agency, and Model Knowledge}
\label{sec:limitations_action}

Support and learning influence one another through the interaction process.
Predictions inform which activities are offered, the person's choices determine which occur, and the resulting responses enter the history used for later prediction.
A policy that repeatedly offers a narrow set of arrangements can therefore accumulate evidence concentrated in those arrangements while leaving alternatives poorly understood.
Historical choices may also depend on unrecorded aspects of the person's condition, so fitting the observed outcomes does not establish what a different choice would produce.
The evidence requirements in Section~\ref{sec:research_assistance} apply to these alternatives and to claims that using the model benefits the person; additional interaction alone does not resolve the gaps.

The person's participation is important both to model revision and to the choice of support.
They may correct an event, challenge an inferred state, or express a new preference, each with different implications for the maintained account.
A factual correction prompts reconsideration of evidence, while a new goal changes the grounds for the current decision without invalidating an earlier preference statement.
Requests for clarification can improve available evidence but also impose effort and change the interaction; participation in updating the model depends on the person's willingness and priorities.
A model may retain an inaccurate or outdated account despite opportunities for correction, and the person's responses and preferences may continue to change.
Its predictions cannot settle what the person should value, and predicted acceptance remains distinct from the person's actual agreement.

\subsection{Open Questions and Future Research}
\label{sec:limitations_open_questions}

These limits motivate several connected research questions:
\begin{itemize}
    \item How can a system recognize when its task-relevant state representation no longer covers the distinctions needed for prediction and support?
    \item What evidence indicates that an earlier personal pattern has become inapplicable, and how should the model adapt while distinguishing personal change from contextual or observational change?
    \item Which historical detail must remain available to revise those judgments?
    \item How can interactions that clarify the model's understanding also respect the person's goals and willingness to participate?
\end{itemize}

Investigating these questions may motivate changes to the representations, learning processes, and interaction assumptions of HOWM.

\section{Conclusion}
\label{sec:conclusion}

We have formulated the Human-Oriented World Model (HOWM) as a conceptual framework centered on a person's task-relevant state and its evolution.
It connects current observations and individual history with environmental and social context to organize continuing state estimation, personalized dynamics prediction, and action support.
The current estimate anchors expectations about possible developments, which an external policy relates to the person's goals, preferences, and constraints when considering support.
Actual responses and outcomes supply experience that can revise the estimate and personal dynamic patterns, changing the basis for the next encounter.

The report develops the modeling objects and mechanisms needed to describe this process.
It also outlines a research agenda for assessing these capabilities and the contributions of different information sources.
The training and learning scenarios illustrate how experience could inform subsequent use, while the limitations identify conditions under which inference, personal patterns, and learning remain uncertain.
The next step is to instantiate HOWM in bounded tasks and test whether accumulated interaction experience improves the tracking and prediction of the same person's evolving state, and whether using that understanding helps the person pursue their goals over time.


\bibliographystyle{plainnat}
\bibliography{Latex_arXiv_HOWM}

\end{document}